\documentclass{article}

\usepackage{spconf}
\usepackage{graphicx}
\usepackage{booktabs}
\usepackage{tabularx}
\usepackage{caption}
\usepackage{array}
\usepackage{colortbl}
\usepackage{xcolor}
\usepackage{makecell}
\usepackage[table]{xcolor}

\usepackage{amsmath, amsfonts, amssymb}
\usepackage{bm}
\newcommand{\methodname}{\textsc{GRPO}}   
\newcommand{\zvec}{\ensuremath{\bm{z}}}

\usepackage{algorithm}
\usepackage{algpseudocode}

\usepackage[T1]{fontenc}
\usepackage[utf8]{inputenc}

\usepackage{tikz}
\usetikzlibrary{positioning, arrows.meta, shapes.geometric, calc}

\newcolumntype{Y}{>{\centering\arraybackslash}X}
\definecolor{bestcolor}{RGB}{255,200,120}

\usepackage{hyperref}

\title{GTMA: Dynamic Representation Optimization for OOD VLMs}
\name{Jensen Zhang \qquad Ningyuan Liu  \qquad Keze Wang}
\address{Sun Yat-sen University}

\begin{document}
\maketitle
\begin{abstract}
Vision-Language Models (VLMs) struggle in open-world applications, where Out-of-Distribution (OOD) concepts often trigger cross-modal alignment collapse and severely degrade zero-shot performance. We identify the root cause as \textbf{Modal Asymmetry}: while the visual encoder can extract discriminative features from unseen images, the text encoder is restricted by a fixed discrete vocabulary and thus cannot synthesize new semantic anchors. Existing approaches such as CoOp or LoRA provide only partial remedies, since they remain confined to the pre-trained semantic space.  
To overcome this bottleneck, we propose \textbf{Dynamic Representation Optimization}, realized through the \textbf{Guided Target-Matching Adaptation (GTMA)} framework. At inference time, GTMA constructs a continuous pseudo-word embedding that best aligns with an OOD image’s visual anchor, thereby bypassing vocabulary limitations. The optimization is driven by our \textbf{Adaptive Gradient-based Representation Policy Optimization (GRPO)} algorithm, which integrates semantic regularization to preserve plausibility and compatibility with the model’s prior knowledge. 
Experiments on ImageNet-R and our VISTA-Beyond benchmark show that GTMA improves zero/few-shot OOD accuracy by up to 15--20\% over the base VLM while maintaining performance on seen concepts. Ablation studies confirm the necessity of pseudo-word optimization. 
\end{abstract}


\section{Introduction}
\label{sec:intro}

Vision-Language Models (VLMs)~\cite{gan2022visionlanguagepretrainingbasicsrecent} are catalyzing a paradigm shift from closed-world recognition toward open-world understanding, powering applications in robotics and autonomous driving~\cite{z1}. However, when deployed in dynamic environments, their generalization capabilities often collapse. This failure stems from encountering \textbf{Out-of-distribution (OOD)}~\cite{z2} novel concepts, which leads to a catastrophic breakdown of internal cross-modal alignment~\cite{kamath2023textencodersbottleneckcompositionality}.

We introduce and formalize the concept of \textbf{``Modal Asymmetry''} to pinpoint the root cause of this phenomenon. For images containing OOD concepts, the visual encoder can still produce highly discriminative representations due to powerful feature transfer~\cite{z3}. In contrast, the text encoder, constrained by its fixed discrete vocabulary, cannot generate effective semantic coordinates for these missing concepts. This text-side bottleneck induces a modal imbalance that is a fundamental barrier to deploying VLMs in the open world.

This problem is difficult to analyze with existing benchmarks like ImageNet-R~\cite{z4}, which possess latent overlap between their label sets and VLM pre-training corpora, failing to offer a truly isolated OOD testbed. To enable precise quantification and rigorous validation, we construct a new comprehensive benchmark—\textbf{VISTA-Beyond}. Through systematic vocabulary screening, VISTA-Beyond is the first to explicitly separate In-Pre-training (IP) and OOD concepts across 16 diverse domains, providing an unprecedented ``clean'' testbed for studying the true OOD generalization of VLMs.

Existing adaptation methods, such as prompt engineering (e.g., CoOp~\cite{z10,z12}) or parameter-efficient fine-tuning (e.g., LoRA~\cite{Z5}), fundamentally reorganize or interpolate among known concepts and cannot synthesize entirely new semantic anchors for unknown concepts. This motivates our key question: can we, at inference time, dynamically synthesize an optimal textual representation for an OOD visual concept by searching within a continuous semantic space?

To this end, we propose a novel test-time adaptation framework named \textbf{Guided Target–Matching Adaptation (GTMA)}. Our approach introduces \textbf{Dynamic Representation Optimization}, which shifts focus from adapting model parameters to directly synthesizing an optimal input representation. For a given OOD visual anchor, GTMA formulates an instance-level optimization task to solve for a continuous pseudo-word embedding that maximally aligns with it. This task is efficiently handled by our \textbf{Adaptive Gradient-based Representation Policy Optimization (GRPO)} algorithm, which integrates semantic regularization to ensure the synthesized embedding remains on a plausible semantic manifold. Our work's main contributions are thus the formal definition of Modal Asymmetry, the creation of the VISTA-Beyond benchmark for its rigorous study, and the proposal of GTMA as an effective solution. Extensive experiments show that GTMA significantly outperforms existing methods, validating its ability to enhance the open-world generalization of VLMs.

\section{Related Work}
\label{sec:relatedwork}

Vision-Language Models (VLMs) like CLIP~\cite{z7} achieve zero-shot capabilities by aligning modalities in a shared semantic space. However, they often fail when encountering out-of-distribution (OOD) concepts. We term this failure \textbf{\textit{Modal Asymmetry}}: the visual encoder can produce discriminative features for novel objects, but the text encoder, constrained by a fixed vocabulary, cannot generate corresponding semantic anchors~\cite{liang2022mindgapunderstandingmodality}. Existing adaptation methods attempt to mitigate this. Parameter-Efficient Fine-Tuning (PEFT) techniques such as LoRA~\cite{DBLP:journals/corr/abs-2106-09685} and prompt learning methods like CoOp~\cite{z6} adapt models with minimal parameter changes. Similarly, Test-Time Adaptation (TTA)~\cite{wang2021tent} updates model components using test data. The fundamental limitation of these approaches is that they primarily adjust model \textit{parameters} and are confined to interpolating within the pre-trained semantic manifold, rendering them unable to represent truly novel concepts. Generative strategies that describe OOD inputs also suffer from vocabulary limitations and error propagation~\cite{DBLP:journals/corr/DaiL15a}.

To overcome these limitations, we propose \textbf{Guided Target–Matching Adaptation (GTMA)}, a framework built on \textbf{\textit{Dynamic Representation Optimization}}. Instead of adapting model parameters, GTMA directly addresses modal asymmetry by synthesizing an optimal, continuous textual representation for each OOD visual input at inference time. This instance-level, input-space optimization is distinct from prior methods like test-time prompt tuning (TPT)~\cite{z8}, which operates on a discrete vocabulary with batch-level objectives. Our approach directly searches for a pseudo-word embedding guided by the visual anchor and regularized by the learned semantic manifold, offering a new pathway to robust open-world generalization.

\begin{figure*}[h!]
  \centering
  \includegraphics[width=0.95\textwidth]{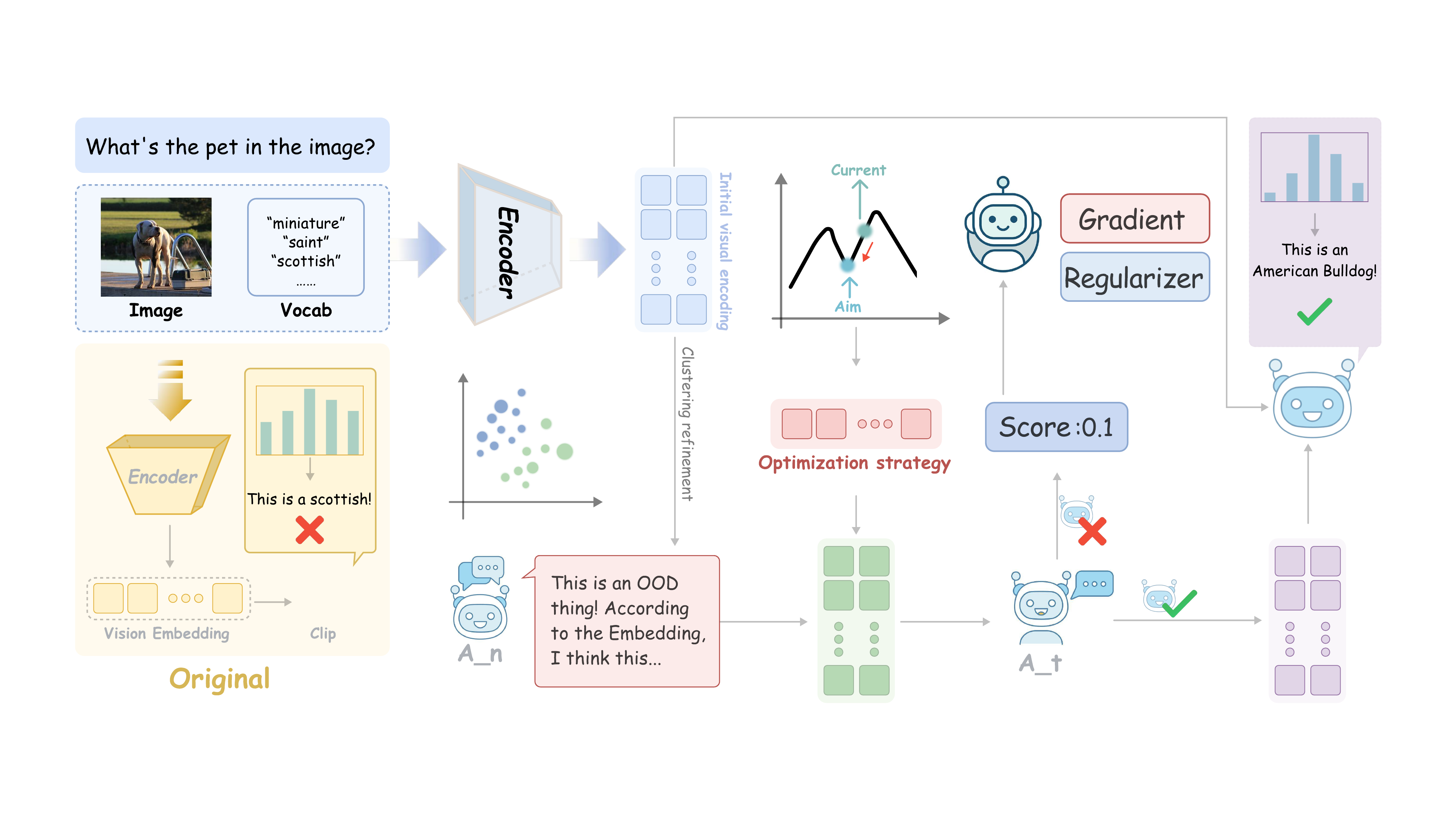}  
  \vspace{-0.5em}
\caption{GTMA inference workflow: for an OOD visual anchor $\mathbf{c}_v$, iteratively optimize a pseudo-word embedding $z$ under visual guidance and semantic regularization, synthesizing textual representations for novel concept recognition.}
  \label{fig:overall_arch}
  \vspace{-1em}
\end{figure*}

\section{Methodology}
\label{sec:methodology}

We introduce \textbf{GTMA (Guided Target-Matching Adaptation)}, a test-time optimization framework designed to address out-of-distribution (OOD) alignment failure in Vision-Language Models (VLMs). We attribute this failure to \textbf{Modal Asymmetry}, where the text encoder's fixed vocabulary creates a bottleneck for representing novel concepts. GTMA circumvents this by dynamically synthesizing a continuous semantic embedding for each OOD instance at inference time.

\subsection{Problem Formulation}
A VLM consists of a visual encoder $\mathcal{V}$ and a text encoder $\mathcal{T}$. For an OOD image $I_{\text{ood}}$, $\mathcal{V}$ can still produce a high-quality, discriminative \textbf{visual anchor}, $\mathbf{c}_v = \mathcal{V}(I_{\text{ood}})$. However, $\mathcal{T}$ cannot natively produce a corresponding embedding. To enhance stability, the raw anchor, derived from patch features $\{\mathbf{f}_i\}$, is purified using a lightweight self-attention mechanism. A global context vector $\bar{\mathbf{f}}$ acts as a query to compute attention weights $\alpha$ over the patches via scaled dot-product attention~\cite{DBLP:journals/corr/VaswaniSPUJGKP17} as $\alpha = \text{softmax}\left(\frac{(\mathbf{W}_q \bar{\mathbf{f}}) (\mathbf{W}_k \mathbf{F})^T}{\sqrt{d_k}}\right)$. The purified anchor is then computed as $\tilde{\mathbf{c}}_v = \sum_{i=1}^{M} \alpha_i \mathbf{f}_i$. For simplicity, we use $\mathbf{c}_v$ to denote this purified anchor henceforth.

\vspace{-4mm}

\subsection{GTMA Framework and GRPO Algorithm}
GTMA recasts OOD alignment as an instance-level optimization problem. The goal is to find an optimal continuous \textbf{pseudo-word embedding} $\mathbf{z} \in \mathbb{R}^d$ that, when injected into a text template $P$ (e.g., ``a photo of a [z]''), maximizes the similarity with the visual anchor $\mathbf{c}_v$, such that $\mathbf{z}^* = \arg\max_{\mathbf{z}} \text{sim}(\mathbf{c}_v, \mathcal{T}(P(\mathbf{z})))$.

We solve this non-convex problem with our \textbf{GRPO (Adaptive Gradient-based Representation Policy Optimization)} algorithm. Starting from an initial guess $\mathbf{z}_0$ (e.g., from an MLP applied to $\mathbf{c}_v$), GRPO iteratively updates the embedding over $T$ steps according to the rule: $\mathbf{z}_{t+1} = \mathbf{z}_t + \eta_t \left( \mathbf{g}_t - \lambda \cdot \nabla R(\mathbf{z}_t) \right)$. The update step is defined by three key components. The \textbf{Gradient ($\mathbf{g}_t$)}, $\nabla_{\mathbf{z}_t} S(\mathbf{z}_t)$, provides the primary optimization direction. \textbf{Semantic Regularization ($R(\mathbf{z}_t)$)}, defined as $R(\mathbf{z}_t) = \frac{1}{2} \|\mathbf{z}_t - \text{Proj}_{E(V)}(\mathbf{z}_t)\|_2^2$, ensures semantic plausibility by penalizing deviation from the manifold of known word embeddings $E(V)$. Finally, the \textbf{Adaptive Learning Rate ($\eta_t$)}, $\eta_t = \eta_0 \cdot \sigma(\beta (\rho_t - \gamma))$, is modulated based on the cosine similarity $\rho_t$ between consecutive gradients to accelerate stable convergence.

\begin{algorithm}[H]
\caption{\textsc{GRPO}: Adaptive Gradient-based Representation Policy Optimization}
\label{alg:grpo}
\renewcommand{\algorithmiccomment}[1]{\hfill\textcolor{gray!80}{\(\triangleright\) #1}}
\begin{algorithmic}[1]
\State \textbf{Input:} Visual anchor $\mathbf{c}_v$, text encoder $\mathcal{T}$, template $P$, iterations $T$
\State \textbf{Hyperparameters:} $\eta_0, \lambda, \beta, \gamma$
\State \textbf{Initialize:} $\mathbf{z}_0 \gets \text{MLP}(\mathbf{c}_v)$; $\mathbf{g}_{-1} \gets \mathbf{0}$
\vspace{0.5em}
\For{$t = 0$ \textbf{to} $T-1$}
    \State $S_t \gets \text{sim}(\mathbf{c}_v, \mathcal{T}(P(\mathbf{z}_t)))$ \Comment{Objective score}
    \State $\mathbf{g}_t \gets \nabla_{\mathbf{z}_t} S_t$ \Comment{Gradient via backprop}
    \State $\nabla R_t \gets \mathbf{z}_t - \text{Proj}_{E(V)}(\mathbf{z}_t)$ \Comment{Reg. gradient}
    \State $\rho_t \gets \text{sim}(\mathbf{g}_t, \mathbf{g}_{t-1})$ \Comment{Gradient similarity}
    \State $\eta_t \gets \eta_0 \cdot \sigma(\beta (\rho_t - \gamma))$ \Comment{Adaptive step size}
    \State $\mathbf{z}_{t+1} \gets \mathbf{z}_t + \eta_t (\mathbf{g}_t - \lambda \nabla R_t)$ \Comment{Update embedding}
    \State $\mathbf{g}_{t-1} \gets \mathbf{g}_t$ \Comment{Track gradient}
\EndFor
\vspace{0.5em}
\State \textbf{Output:} Optimal pseudo-word embedding $\mathbf{z}^* \gets \mathbf{z}_T$
\end{algorithmic}
\end{algorithm}

\begin{table*}[ht]
\centering
\caption{Performance comparison of different methods across six diverse datasets. We report accuracy (\%) under three settings: Out-of-Distribution (OOD), Style Change (SC), and Open-world. The best performance in each column is highlighted in bold.}
\label{tab:main_results}
\resizebox{\textwidth}{!}{%
\begin{tabular}{l ccc ccc ccc ccc ccc ccc}
\toprule
& \multicolumn{3}{c}{\textbf{Texture}} & \multicolumn{3}{c}{\textbf{Satellite images}} & \multicolumn{3}{c}{\textbf{Flowers}} & \multicolumn{3}{c}{\textbf{Pets}} & \multicolumn{3}{c}{\textbf{Cars}} & \multicolumn{3}{c}{\textbf{Action}} \\
\cmidrule(lr){2-4} \cmidrule(lr){5-7} \cmidrule(lr){8-10} \cmidrule(lr){11-13} \cmidrule(lr){14-16} \cmidrule(lr){17-19}
\textbf{Method} & OOD & SC & Open & OOD & SC & Open & OOD & SC & Open & OOD & SC & Open & OOD & SC & Open & OOD & SC & Open \\
\midrule
CoOp & 85.0 & 67.0 & 59.9 & 93.0 & 50.1 & 39.3 & 91.3 & 81.2 & 76.0 & 79.8 & 58.4 & 57.1 & 88.1 & 86.6 & \textbf{85.7} & 79.9 & 68.6 & 65.8 \\
CoCoOp & 84.2 & 67.8 & 60.3 & 92.9 & 56.3 & 41.5 & 90.8 & 85.8 & 81.2 & 82.0 & 74.4 & 73.0 & 88.6 & \textbf{88.4} & 84.5 & 77.6 & 74.7 & 70.0 \\
CLIP-Adapter & 85.9 & \textbf{71.8} & 62.4 & 93.0 & 66.8 & 53.7 & 91.8 & 87.8 & 83.4 & \textbf{91.4} & \textbf{89.2} & 54.2 & 88.5 & 85.7 & 85.5 & 81.0 & \textbf{80.6} & \textbf{73.5} \\
LAION-Beyond & 83.1 & 71.2 & 62.9 & 81.4 & \textbf{67.1} & 54.5 & 90.1 & 86.8 & 83.0 & 90.7 & 66.3 & 64.3 & \textbf{88.7} & 85.1 & \textbf{85.7} & 81.2 & 73.5 & 69.1 \\
\midrule
\rowcolor{gray!25} 
\textbf{GTMA (Ours)} & \textbf{86.1} & 71.7 & \textbf{64.3} & \textbf{93.5} & 66.9 & \textbf{55.4} & \textbf{93.0} & \textbf{89.9} & \textbf{88.2} & 91.0 & 86.8 & \textbf{84.6} & \textbf{88.7} & 88.3 & 84.9 & \textbf{81.9} & 74.7 & 70.2 \\
\bottomrule
\end{tabular}%
}
\end{table*}

\section{Experiments}
\label{sec:experiment}
In this section, we present a comprehensive experimental evaluation of our proposed framework, \textbf{GTMA (Guided Target-Matching Adaptation)}. We detail the experimental methodology, hyperparameter settings, and evaluation results under both Few-Shot and Zero-Shot~\cite{5206594} settings on a diverse set of challenging benchmarks. Compared to previous baseline models, GTMA demonstrates a significant improvement in image classification accuracy in both scenarios, validating the feasibility of our dynamic representation optimization approach.

\subsection{Experimental Settings}

To evaluate our \textbf{GTMA} framework, we compare it against state-of-the-art methods like \textbf{CoOp}~\cite{DBLP:journals/corr/abs-2109-01134}, \textbf{CoCoOp}~\cite{Zhou_2022_CVPR}, and \textbf{CLIP-Adapter}~\cite{chen2025laionbeyond}, using \textbf{OpenCLIP}~\cite{Cherti_2023} as the backbone for all models to ensure a fair comparison. Our GRPO algorithm was configured with $T=10$ optimization iterations, an initial learning rate $\eta_0=0.01$, and a regularization weight $\lambda_{\text{reg}}=0.1$. All evaluations were conducted on our custom \textbf{VISTA-Beyond} benchmark ~\cite{CZKL}, which is meticulously curated from LAION-400M/2B to explicitly separate \textbf{Seen-Concept (SC)} and \textbf{Out-of-Distribution (OOD)} classes. The primary evaluation metric is classification \textbf{accuracy}. For the few-shot experiments, we fine-tune all models on 1, 2, 4, 8, and 16 OOD samples using the OpenCLIP ViT-B/16 model.

\subsection{Few-Shot Learning}
In this experiment, we evaluate performance by fine-tuning the OpenCLIP ViT-B/16 backbone on 1, 2, 4, 8, and 16 Out-of-Distribution (OOD) samples. To test open-vocabulary capabilities, models are trained only on OOD data and subsequently evaluated on a combined test set of both OOD and Seen Concept (SC) images.

\begin{figure}[h!]
\begin{center}
\centerline{\includegraphics[width=\linewidth]{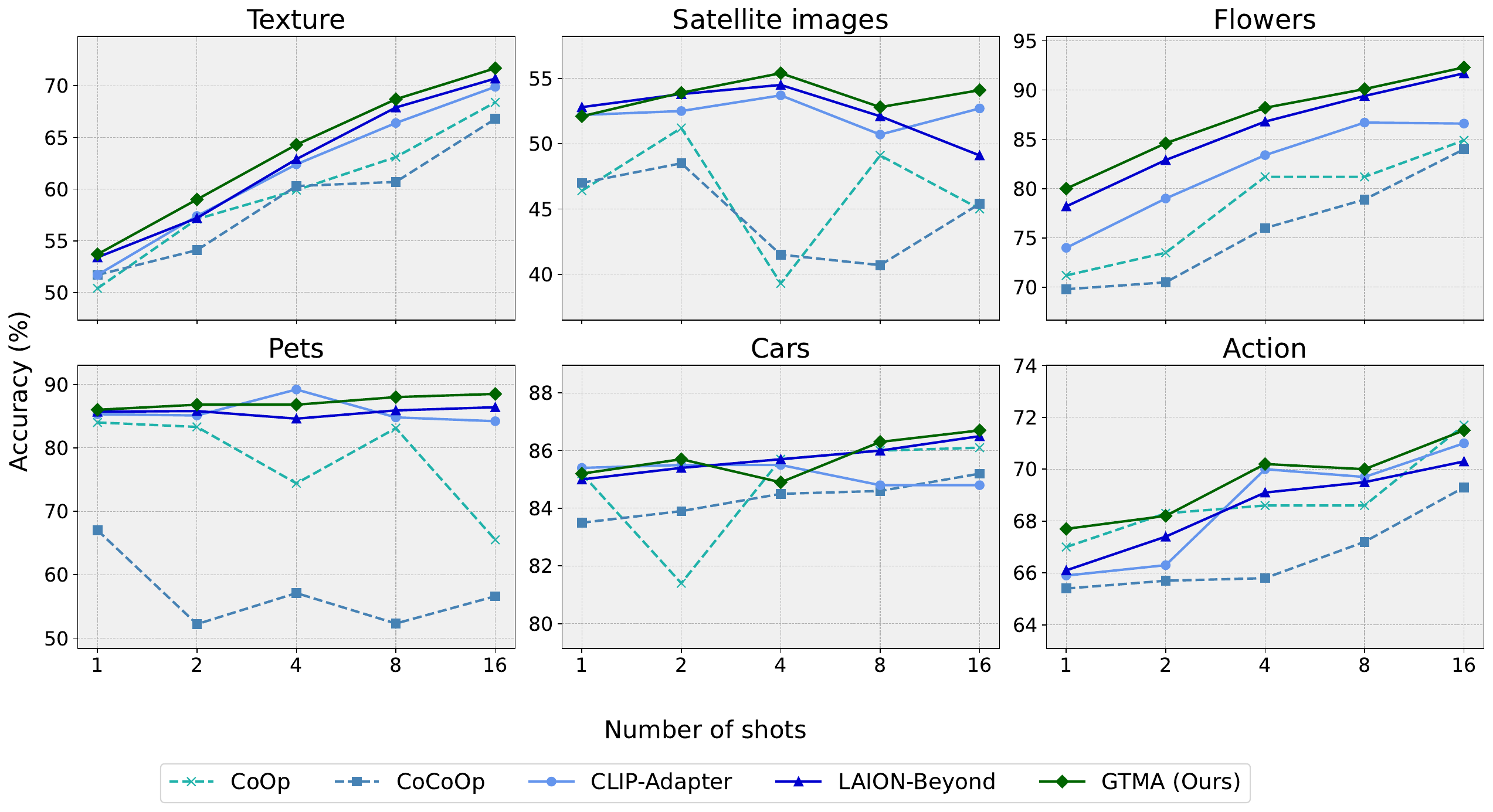}}
\caption{Accuracy scales positively with the number of shots. Across all settings, GTMA maintains a consistent accuracy advantage over baselines due to its dynamic representation optimization.}
\label{fig:performance_shot}
\end{center}
\vskip -0.3in
\end{figure}

As shown in Table~\ref{tab:main_results}, the results highlight GTMA's superior performance on both SC and OOD concepts. This stems from its novel optimization framework, which dynamically synthesizes an optimal pseudo-word embedding ($\mathbf{z}^*$) for each OOD visual concept to directly address modal asymmetry. Concurrently, a semantic regularization term, $R(\mathbf{z}_t) = \frac{1}{2} \|\mathbf{z}_t - \text{Proj}_{E(V)}(\mathbf{z}_t)\|_2^2$, anchors the optimization near the manifold of known concepts, which prevents catastrophic forgetting of SC knowledge. This dual-action approach enables GTMA to effectively balance generalization and retention, achieving leading scores on datasets like Flowers (88.2\%) and Texture (64.3\%). As depicted in Figure~\ref{fig:performance_shot}, the accuracy of all methods improves as the number of training shots increases.

\subsection{Zero-Shot Learning}

To simulate a true zero-shot scenario, we follow the protocol of prior work, using a dataset that provides training samples for OOD categories but masking their labels. This ensures the model's evaluation relies solely on its ability to generalize from the provided images, not on any direct supervision for the target classes. A total of 16 train-test domain splits are constructed to rigorously validate Zero-Shot performance. We employ OpenCLIP, TransCLIP~\cite{zanella2024boostingvisionlanguagemodelstransduction} and LAION-Beyond as our baseline.

\begin{table}[h!]
\centering
\caption{Zero-shot model accuracy (\%) across various datasets.}
\label{tab:zeroshot_acc}
\renewcommand{\arraystretch}{1.2}
\setlength{\tabcolsep}{6pt}
\resizebox{0.95\linewidth}{!}{%
\begin{tabular}{l|cccc}
\toprule
\rowcolor{gray!15}
\textbf{Dataset} & \textbf{OpenCLIP} & \textbf{TransCLIP} & \textbf{LAION-Beyond} & \textbf{GTMA (Ours)} \\
\midrule
Texture          & 48.5 & 50.2 & 55.8 & \textbf{56.1} \\
Satellite Images & 45.1 & 46.3 & 50.1 & \textbf{50.7} \\
Flowers          & 66.1 & 67.5 & 72.3 & \textbf{73.1} \\
Pets             & 85.2 & 86.1 & 88.9 & \textbf{89.2} \\
Cars             & 85.0 & 86.4 & 90.5 & \textbf{91.6} \\
Action           & 66.3 & 67.8 & 71.2 & \textbf{72.9} \\
\bottomrule
\end{tabular}%
}
\end{table}

The zero-shot results, presented in Table~\ref{tab:zeroshot_acc}, confirm the effectiveness of our approach even without any target-domain examples. This superior zero-shot capability is attributed to GTMA's core mechanism. At inference time, the framework can synthesize a novel semantic anchor ($\mathbf{z}^*$) for an unseen visual concept ($\mathbf{c}_v$) purely by optimizing the alignment objective, without requiring any pre-existing textual labels for the new concept. This ability to create semantic representations on-the-fly is what enables strong generalization even without direct supervision.

\subsection{Ablation Study}
To dissect the contributions of our framework's components, we conducted an ablation study on the VISTA-Beyond benchmark. As summarized in Table~\ref{tab:ablation}, the results highlight the integral role each element plays.

\begin{table}[h!]
\centering
\caption{GTMA component ablation: classification accuracy (\%) and mean drop across three datasets per component.}
\label{tab:ablation}
\renewcommand{\arraystretch}{1.2}
\setlength{\tabcolsep}{6pt}
\resizebox{0.9\linewidth}{!}{%
\begin{tabular}{l|ccc|c}
\toprule
\rowcolor{gray!15}
\textbf{Model Variant} & \textbf{Texture} & \textbf{Satellite} & \textbf{Flowers} & \textbf{Avg. Drop} \\
\midrule
\textbf{Full GTMA (Ours)} & \textbf{64.3\%} & \textbf{55.4\%} & \textbf{88.2\%} & \textbf{--} \\
\midrule
w/o Pseudo-word Optimization & 60.3\% & 51.3\% & 68.3\% & -9.3\% \\
w/o Visual Anchor Refinement & 60.8\% & 51.4\% & 68.0\% & -9.2\% \\
w/o Adaptive Learning Rate & 61.1\% & 51.8\% & 68.5\% & -8.8\% \\
w/o Semantic Regularization & 61.8\% & 52.6\% & 69.6\% & -8.0\% \\
\bottomrule
\end{tabular}%
}
\end{table}

Our analysis reveals that each component is critical. The most significant performance drop of \textbf{-9.3\%} occurs without \textbf{Pseudo-word Optimization}, confirming that on-the-fly semantic synthesis is essential for bridging the OOD modality gap. Similarly, removing \textbf{Visual Anchor Refinement} results in a \textbf{-9.2\%} decline, highlighting its importance in guiding the optimization toward meaningful semantics. Finally, the substantial drops from removing the \textbf{Adaptive Learning Rate} (\textbf{-8.8\%}) and \textbf{Semantic Regularization} (\textbf{-8.0\%}) underscore their indispensable roles in stabilizing and constraining the search for the optimal textual representation.

\section{Conclusion}
\label{sec:conclusion}
In this paper, we identify \textbf{Modal Asymmetry} as the root cause of OOD failures in Vision-Language Models (VLMs) and introduce the \textbf{VISTA-Beyond} benchmark for its rigorous study. To address this, we propose \textbf{\methodname{}}, a framework with a novel \textbf{Dynamic Representation Optimization} strategy. By synthesizing continuous pseudo-word embeddings ($\zvec^*$) at inference, \methodname{} mitigates the fixed vocabulary bottleneck. Experiments show that \methodname{} outperforms baselines in zero-/few-shot OOD generalization while preserving seen-concept performance. Our work points to a promising path toward more robust VLMs for open-world applications, with future potential in open-vocabulary detection.ection.

\bibliographystyle{IEEEbib}
\bibliography{strings}

@article{DBLP:journals/corr/abs-2109-01134,
  author       = {Kaiyang Zhou and
                  Jingkang Yang and
                  Chen Change Loy and
                  Ziwei Liu},
  title        = {Learning to Prompt for Vision-Language Models},
  journal      = {CoRR},
  volume       = {abs/2109.01134},
  year         = {2021},
  url          = {https://arxiv.org/abs/2109.01134},
  eprinttype    = {arXiv},
  eprint       = {2109.01134},
  bibsource    = {dblp computer science bibliography, https://dblp.org}
}

@InProceedings{Zhou_2022_CVPR,
    author    = {Zhou, Kaiyang and Yang, Jingkang and Loy, Chen Change and Liu, Ziwei},
    title     = {Conditional Prompt Learning for Vision-Language Models},
    booktitle = {Proceedings of the IEEE/CVF Conference on Computer Vision and Pattern Recognition (CVPR)},
    month     = {June},
    year      = {2022},
    pages     = {16816-16825}
}

@article{DBLP:journals/corr/abs-2106-09685,
  author       = {Edward J. Hu and
                  Yelong Shen and
                  Phillip Wallis and
                  Zeyuan Allen{-}Zhu and
                  Yuanzhi Li and
                  Shean Wang and
                  Weizhu Chen},
  title        = {LoRA: Low-Rank Adaptation of Large Language Models},
  journal      = {CoRR},
  volume       = {abs/2106.09685},
  year         = {2021},
  url          = {https://arxiv.org/abs/2106.09685},
  eprinttype    = {arXiv},
  eprint       = {2106.09685},
  bibsource    = {dblp computer science bibliography, https://dblp.org}
}

@inproceedings{
wang2021tent,
title={Tent: Fully Test-Time Adaptation by Entropy Minimization},
author={Dequan Wang and Evan Shelhamer and Shaoteng Liu and Bruno Olshausen and Trevor Darrell},
booktitle={International Conference on Learning Representations},
year={2021},
url={https://openreview.net/forum?id=uXl3bZLkr3c}
}

@article{DBLP:journals/corr/DaiL15a,
  author       = {Andrew M. Dai and
                  Quoc V. Le},
  title        = {Semi-supervised Sequence Learning},
  journal      = {CoRR},
  volume       = {abs/1511.01432},
  year         = {2015},
  url          = {http://arxiv.org/abs/1511.01432},
  eprinttype    = {arXiv},
  eprint       = {1511.01432},
  bibsource    = {dblp computer science bibliography, https://dblp.org}
}

@article{DBLP:journals/corr/VaswaniSPUJGKP17,
  author       = {Ashish Vaswani and
                  Noam Shazeer and
                  Niki Parmar },
  title        = {Attention Is All You Need},
  journal      = {CoRR},
  volume       = {abs/1706.03762},
  year         = {2017},
  url          = {http://arxiv.org/abs/1706.03762},
  eprinttype    = {arXiv},
  eprint       = {1706.03762},
  bibsource    = {dblp computer science bibliography, https://dblp.org}
}

@misc{liang2022mindgapunderstandingmodality,
      title={Mind the Gap: Understanding the Modality Gap in Multi-modal Contrastive Representation Learning}, 
      author={Weixin Liang and Yuhui Zhang and Yongchan Kwon and Serena Yeung and James Zou},
      year={2022},
      eprint={2203.02053},
      archivePrefix={arXiv},
      primaryClass={cs.CL},
      url={https://arxiv.org/abs/2203.02053}, 
}

@misc{gan2022visionlanguagepretrainingbasicsrecent,
      title={Vision-Language Pre-training: Basics, Recent Advances, and Future Trends}, 
      author={Zhe Gan and Linjie Li and Chunyuan Li and Lijuan Wang and Zicheng Liu and Jianfeng Gao},
      year={2022},
      eprint={2210.09263},
      archivePrefix={arXiv},
      primaryClass={cs.CV},
      url={https://arxiv.org/abs/2210.09263}, 
}

@INPROCEEDINGS{5206594,
  author={Lampert, Christoph H. and Nickisch, Hannes and Harmeling, Stefan},
  booktitle={2009 IEEE Conference on Computer Vision and Pattern Recognition}, 
  title={Learning to detect unseen object classes by between-class attribute transfer}, 
  year={2009},
  volume={},
  number={},
  pages={951-958},
  doi={10.1109/CVPR.2009.5206594}}

@inproceedings{chen2025laionbeyond,
    title={Reproducible Vision-Language Models Meet Concepts Out of Pre-Training},
    author={Chen, Ziliang and Huang, Xin },
    booktitle={Proceedings of the IEEE/CVF Conference on Computer Vision and Pattern Recognition (CVPR)},
    pages={xxxx--xxxx},
    year={2025}
  }

@misc{zanella2024boostingvisionlanguagemodelstransduction,
      title={Boosting Vision-Language Models with Transduction}, 
      author={Maxime Zanella and Benoît Gérin and Ismail Ben Ayed},
      year={2024},
      eprint={2406.01837},
      archivePrefix={arXiv},
      primaryClass={cs.CV},
      url={https://arxiv.org/abs/2406.01837}, 
}

@misc{kamath2023textencodersbottleneckcompositionality,
      title={Text encoders bottleneck compositionality in contrastive vision-language models}, 
      author={Amita Kamath and Jack Hessel and Kai-Wei Chang},
      year={2023},
      eprint={2305.14897},
      archivePrefix={arXiv},
      primaryClass={cs.CL},
      url={https://arxiv.org/abs/2305.14897}, 
}

@INPROCEEDINGS{CZKL,
  author={Chen, Ziliang and Huang, Xin and Fan, Xiaoxuan and Wang, Keze and Zhou, Yuyu and Guan, Quanlong and Lin, Liang},
  booktitle={2025 IEEE/CVF Conference on Computer Vision and Pattern Recognition (CVPR)}, 
  title={Reproducible Vision-Language Models Meet Concepts Out of Pre-Training}, 
  year={2025},
  volume={},
  number={},
  pages={14701-14711},
  doi={10.1109/CVPR52734.2025.01370}}

@inproceedings{
z1,
title={{KABB}: Knowledge-Aware Bayesian Bandits for Dynamic Expert Coordination in Multi-Agent Systems},
author={Jusheng Zhang and Zimeng Huang and Yijia Fan and Ningyuan Liu and Mingyan Li and Zhuojie Yang and Jiawei Yao and Jian Wang and Keze Wang},
booktitle={Forty-second International Conference on Machine Learning},
year={2025},
url={https://openreview.net/forum?id=AKvy9a4jho}
}

@inproceedings{
z2,
title={{GAM}-Agent: Game-Theoretic and Uncertainty-Aware Collaboration for Complex Visual Reasoning},
author={Jusheng Zhang and Yijia Fan and Wenjun Lin and Ruiqi Chen and Haoyi Jiang and Wenhao Chai and Jian Wang and Keze Wang},
booktitle={The Thirty-ninth Annual Conference on Neural Information Processing Systems},
year={2025},
url={https://openreview.net/forum?id=EKJhU5ioSo}
}

@misc{z3,
      title={CF-VLM:CounterFactual Vision-Language Fine-tuning}, 
      author={Jusheng Zhang and Kaitong Cai and Yijia Fan and Jian Wang and Keze Wang},
      year={2025},
      eprint={2506.17267},
      archivePrefix={arXiv},
      primaryClass={cs.LG},
      url={https://arxiv.org/abs/2506.17267}, 
}

@inproceedings{
z4,
title={{MAT}-Agent: Adaptive Multi-Agent Training Optimization},
author={Jusheng Zhang and Kaitong Cai and Yijia Fan and Ningyuan Liu and Keze Wang},
booktitle={The Thirty-ninth Annual Conference on Neural Information Processing Systems},
year={2025},
url={https://openreview.net/forum?id=YDWRTYgR79}
}

@inproceedings{
Z5,
title={Tri-{MARF}: A Tri-Modal Multi-Agent Responsive Framework for Comprehensive 3D Object Annotation},
author={Jusheng Zhang and Yijia Fan and Zimo Wen and Jian Wang and Keze Wang},
booktitle={The Thirty-ninth Annual Conference on Neural Information Processing Systems},
year={2025},
url={https://openreview.net/forum?id=YGIbwfNWot}
}

@misc{z6,
      title={MM-CoT:A Benchmark for Probing Visual Chain-of-Thought Reasoning in Multimodal Models}, 
      author={Jusheng Zhang and Kaitong Cai and Xiaoyang Guo and Sidi Liu and Qinhan Lv and Ruiqi Chen and Jing Yang and Yijia Fan and Xiaofei Sun and Jian Wang and Ziliang Chen and Liang Lin and Keze Wang},
      year={2025},
      eprint={2512.08228},
      archivePrefix={arXiv},
      primaryClass={cs.CV},
      url={https://arxiv.org/abs/2512.08228}, 
}

@misc{z7,
      title={HybridToken-VLM: Hybrid Token Compression for Vision-Language Models}, 
      author={Jusheng Zhang and Xiaoyang Guo and Kaitong Cai and Qinhan Lv and Yijia Fan and Wenhao Chai and Jian Wang and Keze Wang},
      year={2025},
      eprint={2512.08240},
      archivePrefix={arXiv},
      primaryClass={cs.CV},
      url={https://arxiv.org/abs/2512.08240}, 
}

@misc{z8,
      title={Failure-Driven Workflow Refinement}, 
      author={Jusheng Zhang and Kaitong Cai and Qinglin Zeng and Ningyuan Liu and Stephen Fan and Ziliang Chen and Keze Wang},
      year={2025},
      eprint={2510.10035},
      archivePrefix={arXiv},
      primaryClass={cs.AI},
      url={https://arxiv.org/abs/2510.10035}, 
}

@misc{z10,
      title={Learning Dynamics of VLM Finetuning}, 
      author={Jusheng Zhang and Kaitong Cai and Jing Yang and Keze Wang},
      year={2025},
      eprint={2510.11978},
      archivePrefix={arXiv},
      primaryClass={cs.LG},
      url={https://arxiv.org/abs/2510.11978}, 
}

@misc{z12,
      title={Kolmogorov-Arnold Fourier Networks}, 
      author={Jusheng Zhang and Yijia Fan and Kaitong Cai and Keze Wang},
      year={2025},
      eprint={2502.06018},
      archivePrefix={arXiv},
      primaryClass={cs.LG},
      url={https://arxiv.org/abs/2502.06018}, 
}

\end{document}